\documentclass[]{SPAICE}

\def\authorEmail{oscar.alcarria@uclm.es}
\def\rafaelEmail{rafael.sanchez@uclm.es}
\def\AuthorShort{Ó. Alcarria, R. Sánchez et al.}

\author[1]{Óscar Alcarria\thanks{Corresponding author. E-Mail: \authorEmail}}
\author[1]{Rafael Sánchez\thanks{Corresponding author. E-Mail: \rafaelEmail}}
\author[2]{Javier Sempere}
\author[1]{Pablo Torrijos}

\author[1]{Juan C. Alfaro}
\author[3]{Juan M. Auñón}
\author[1]{José A. Gámez}
\author[1]{José M. Puerta}

\affil[1]{Universidad de Castilla-La Mancha, Albacete, Spain}
\affil[2]{GMV GmbH, Darmstadt, Germany}
\affil[3]{GMV, Madrid, Spain}

\title{Structure-Aware Unsupervised Anomaly Detection for Spacecraft Telemetry with Adaptive EVT Thresholding}

\begin{document}

\maketitle

\begin{abstract}

Operational anomaly detection in spacecraft telemetry typically requires labeled historical anomalies or extended warm-up periods. These requirements are rarely met in practice. We propose an unsupervised, deployment-ready framework that produces predictions from the second month of operation without any labels, prior fault knowledge, or mission-specific tuning. The approach combines incremental monthly retraining, statistical model selection, and adaptive Extreme Value Theory (EVT) thresholding for false alarm control. On the ESA Anomalies Dataset (ESA-AD), it achieves $F_{0.5}=0.700$ on Mission~1 and $F_{0.5}=0.698$ on Mission~2 under strict chronological evaluation.
\end{abstract}

\sloppy

\section{Introduction}

Anomaly detection in spacecraft telemetry involves identifying off-nominal behavior in multivariate, non-stationary time series under strict operational constraints. 
In practice, mission control environments rarely provide complete labeled fault histories, dedicated GPU hardware, or years of accumulated telemetry before monitoring must begin.

The European Space Agency (ESA) Anomalies Dataset (ESA-AD)~\cite{ESA-ADB-Kotowski2025} is the most comprehensive publicly available benchmark for anomaly detection in spacecraft telemetry, providing expert-annotated data from three complete ESA missions with a hierarchical event-wise evaluation pipeline. Two missions are used for benchmarking: Mission~1 (M1) features 76~telemetry channels (58~target) with 118~anomalies and 78~rare nominal events; Mission~2 (M2) features 100~channels (47~target) with 31~anomalies and 613~rare nominal events.

An exploratory analysis of the ESA-AD reveals marked statistical differences across missions, suggesting that a single modeling strategy may not generalize consistently. We propose a structure-aware framework that: (1)~selects the detection paradigm based on mission-level statistical properties; (2)~integrates incremental monthly retraining for deployment from the second month of operation; and (3)~applies adaptive Extreme Value Theory (EVT)-based thresholding for robust, precision-oriented event-wise anomaly detection. Without labels, GPU hardware, or extensive historical data, the framework matches or surpasses industrial supervised baselines~on~ESA-AD.



\section{Methodology}

\subsection{Problem Definition and Constraints}
We address the detection of anomalous events in multivariate satellite telemetry under fully anonymized conditions, preventing the use of mission-specific semantics or physics-based priors. Following the ESA Benchmark for Anomaly Detection in Satellite Telemetry (ESA-ADB) \cite{ESA-ADB-Kotowski2025} definition, anomalies are modeled as contiguous off-nominal temporal segments rather than isolated~outliers.
To reflect realistic deployment constraints \cite{tejedor2025}, our framework operates without labels or GPU hardware and is evaluated under a chronological incremental setup that preserves temporal causality and avoids information leakage.
Performance is measured using the corrected event-wise $F_{0.5}$ score \cite{ESA-ADB-Kotowski2025}, where $\beta=0.5$ weights precision higher than recall and the event-wise precision is corrected by the sample-level true negative rate to penalize algorithms that over-flag nominal~periods.

\begin{figure*}[ht]
        \centering
        \includegraphics[width=0.87\textwidth]{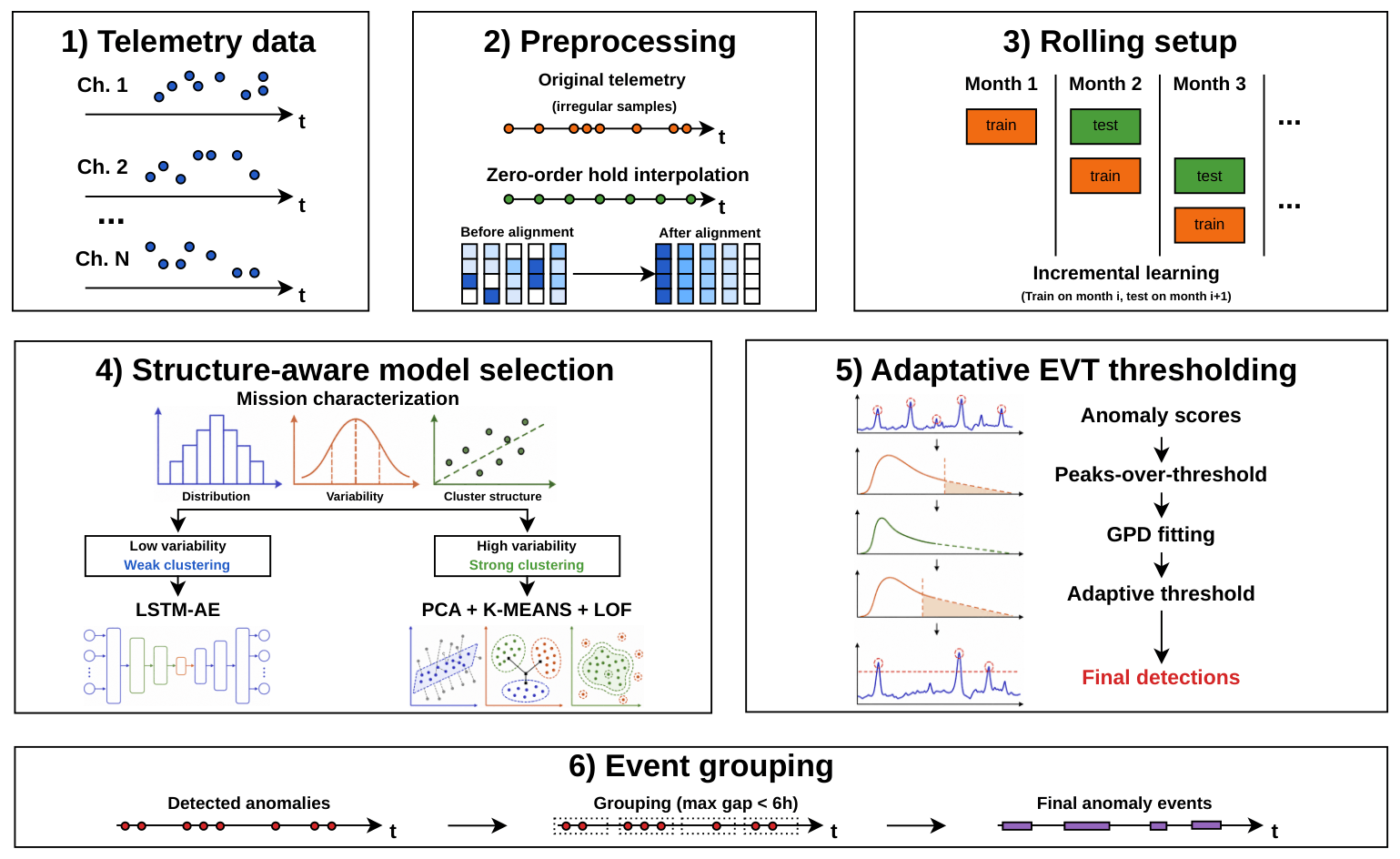}
	\caption{Overview of the proposed structure-aware anomaly detection framework.}
	\label{fig:methodology-overview}
\end{figure*}

Figure~\ref{fig:methodology-overview} summarizes the proposed workflow, from preprocessing and monthly incremental learning to structure-aware model selection, adaptive EVT thresholding, and post-detection anomaly grouping.

\subsection{Data Preprocessing}
Following the preprocessing protocol in~\cite{tejedor2025}, the asynchronous and irregularly sampled telemetry channels are first combined over their original timestamps, producing a sparse multivariate time series. Each channel is then processed independently: missing entries are discarded and the remaining observations are projected onto a mission-specific uniform grid with sampling intervals of 30 s for M1 and 18 s for M2. Zero-order hold interpolation is applied, such that each grid timestamp receives the most recently observed channel value.
To prevent short annotated events from disappearing during temporal discretization, a channel-aware preservation procedure is applied after resampling. For each grid interval overlapping an event annotation, the last original telemetry value recorded within the annotated interval is copied to the following grid timestamp. The resampled channels are subsequently aligned on the common temporal grid and concatenated into a single multivariate series. Finally, event labels are assigned using inclusive start and end times, with samples encoded as normal (\(y=0\)), rare event (\(y=1\)), or anomaly (\(y=2\)); anomalies take precedence when annotated intervals overlap. The resulting dataset is verified to contain a regular datetime index and no missing values.

\subsection{Incremental Learning Setup}
Instead of relying on the static partitioning from ESA-ADB, we implement an incremental learning strategy utilizing a monthly sliding window. In this setup, the model is trained exclusively on the previous month's data and evaluated on the subsequent month, allowing for deployment from the very first available period without the need for large historical datasets. This strategy is well suited to operational environments, as it allows the detector to adapt progressively to changes in telemetry behavior over time and aligns with the update cycles typically found in production monitoring systems.

\subsection{Mission Characterization}
\label{3.4: Mission characterization}
To characterize the missions and guide the selection of appropriate anomaly detection models, we conducted a comparative statistical analysis on the complete datasets of both missions from the ESA-ADB benchmark, including nominal, rare, and anomalous events.

First, the Energy Distance \cite{EnergyDistance-Szekely2013} between M1 and M2 is $4.33$, indicating substantial multivariate distributional differences. 
Statistical significance was assessed via a permutation test (200 permutations) \cite{PermutationTest-Good2005}, in which mission labels were randomly reassigned while preserving sample sizes. 
No permuted statistic exceeded the observed value ($p < 0.005$), confirming the absence of a shared statistical structure between missions. For each mission, the coefficient of variation (CV) was computed independently for each sensor as the ratio of its standard deviation to its absolute mean, and subsequently averaged across all numerical variables. This analysis reveals low variability in M1 (CV$ = 0.13$), consistent with relatively stable dynamics, whereas M2 exhibits markedly higher variability (CV$ = 1.27$), reflecting greater overall dispersion \mbox{across~sensors.}

\begin{figure}[ht]
        \centering
        \includegraphics[width=\columnwidth]{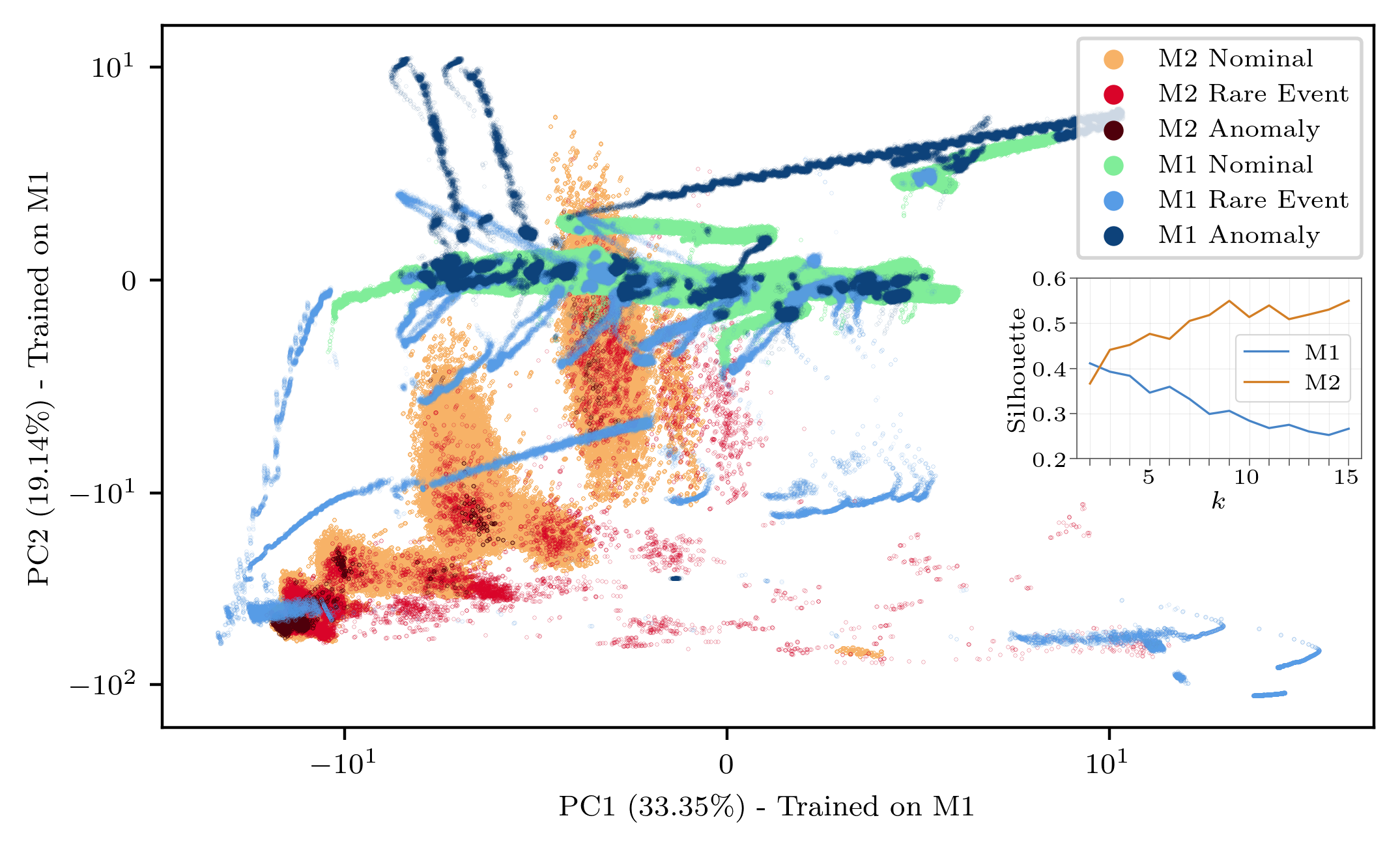}
	\caption{PCA projection of Mission 1 (M1) and Mission 2 (M2). The model is fitted exclusively on M1 using all its variables, which are common to M2. The inset shows the Silhouette score as a function of the number of clusters ($k$).}
	\label{fig:pca}
\end{figure}

Clustering structure was evaluated using K-means \cite{KMeans-MacQueen1967} with Euclidean distance on standardized features, exploring multiple values of $k$. The Silhouette coefficient \cite{Silhouette-Rousseeuw1987} (inset in Fig. \ref{fig:pca}) yields consistently higher scores for M2, suggesting more compact and well-separated clusters, while M1 displays weaker intrinsic segmentation. Principal Component Analysis (PCA) projections \cite{PCA-Hotelling1933} (Fig. \ref{fig:pca}) obtained by fitting PCA on M1 and applying the transformation to both missions, further reinforce this distinction. The first two principal components explain 33.35\% and 19.14\% of the total variance, respectively. In this shared subspace, M1 presents a largely continuous and overlapping distribution structured around a limited number of principal linear directions, whereas M2 exhibits more separated dense regions and isolated low-density areas, consistent with stronger local clustering structure and variance distributed across \mbox{multiple~dimensions.}


Overall, these observations motivate the hypothesis that in M1 anomalous behavior may manifest as subtle deviations embedded within stable dynamics, whereas in M2 anomalies may align with clearer \mbox{structural~separations.} The statistical analysis rules out a single-model approach and motivates a structure-driven selection. Generic standalone models (e.g. single local-density detector or autoencoder) performed poorly in preliminary experiments, confirming that tailored strategies are required to accommodate mission variability.

\subsection{Model Selection and Training}
The model selection strategy is directly informed by the statistical characterization presented in the previous phase. Since the two missions exhibit fundamentally different statistical structures, distinct modeling approaches are adopted within the same methodological framework.

\subsubsection{Temporal Reconstruction Model}
M1 is characterized by low variability and a weak intrinsic clustering structure. These properties suggest that anomalies are likely to manifest as subtle temporal deviations embedded within otherwise stable signals.
Accordingly, an unsupervised Long Short-Term Memory (LSTM) autoencoder \cite{LSTM-AE-Malhotra2016} is employed. This architecture enables the capture of both temporal dynamics and nonlinear relationships among variables, which is essential in scenarios characterized by stable signals and low-magnitude anomalies. For each input sequence $\mathbf{x}_t \in \mathbb{R}^{L \times F}$, where $L$ is the sequence length and $F$ is the number of features, the model reconstructs the input~as~$\hat{\mathbf{x}}_t$. 

To comply with CPU-only constraints and monthly incremental retraining, a lightweight LSTM autoencoder is adopted. Sequences of length $L = 64$ are processed by a single-layer architecture with 64 hidden units and a 16-dimensional latent space, limiting model capacity to reduce overfitting and ensure efficient retraining. The model is optimized with Adam (learning rate $10^{-3}$) for up to 30 epochs, with a batch size of 128. Early stopping (patience = 3) monitors the training reconstruction loss and triggers in most monthly windows, indicating that convergence typically occurs before the epoch limit. Prior to each training cycle, robust sample-level filtering based on the Median Absolute Deviation (MAD) is applied to reduce the influence of outliers on the learned reconstruction patterns.

The reconstruction error for each sequence is quantified as mean squared error (MSE):
\vspace{-0.1cm}
\begin{equation}
s_t = \frac{1}{L F} \sum_{i=1}^{L} \sum_{j=1}^{F} \left( x_{t-L+i,\,j} - \hat{x}_{t-L+i,\,j} \right)^2
\end{equation}
\vspace{-0.2cm}

The resulting anomaly scores $\{s_t\}$ may shift across temporal windows due to non-stationary system dynamics, making fixed thresholds unreliable. An initial threshold is set at the 98th percentile of the training reconstruction error and subsequently refined through the adaptive EVT-based procedure described in Section~\ref{3.6: Dynamic Thresholding}.

\subsubsection{Clustering-Based Framework}
M2 exhibits high variability, pronounced clustering structure, and variance distributed across multiple dimensions. These characteristics indicate the presence of well-defined operational regimes and locally structured state spaces. To exploit this structure, a clustering-based anomaly detection framework is adopted. The pipeline consists of: 
(1) Robust scaling and preliminary unsupervised filtering of extreme outliers as in the previous model; 
(2) PCA-based dimensionality reduction to learn a compact representation of nominal behavior; 
(3) K-means clustering to model operational regimes in the reduced subspace; 
(4) Local Outlier Factor (LOF) \cite{LOF-Breunig2000} to quantify local density deviations; and 
(5) a combined anomaly score integrating regime proximity and local density information.

PCA reduces dimensionality while preserving the dominant structure of normal operation. K-means captures recurrent operational modes, and LOF evaluates the degree of local deviation within each regime. The anomaly score combines the LOF novelty score with the distance to the nearest cluster centroid in the PCA subspace, giving LOF (0.7) a higher weight than centroid distance (0.3), thereby prioritizing local density distortions while retaining sensitivity to global regime shifts.

Hyperparameters are automatically inferred from each training window without supervised tuning. PCA dimensionality is set to explain 95\% cumulative variance. The number of clusters is estimated using the Gap Statistic \cite{GapStatistic-Tibshirani2002} within a data-dependent range, and the LOF neighborhood size scales with $\sqrt{n}$ and the reduced dimensionality to ensure stable density estimation. Anomaly scores are first calibrated using the 98th percentile before applying the dynamic thresholding procedure described in Section~\ref{3.6: Dynamic Thresholding}.

\subsection{EVT-Based Dynamic Thresholding}
\label{3.6: Dynamic Thresholding}

Because anomaly-score distributions may vary across temporal windows, fixed thresholds can become unreliable. We therefore apply a peaks-over-threshold EVT procedure~\cite{EVT-Siffer2017}, which adapts the threshold to the upper tail of the calibration-score distribution.

For window $k$, let $S_k=\{s_{k,1},\ldots,s_{k,n_k}\}$ denote the calibration scores and define the preliminary threshold as $u_k=Q_q(S_k)$, where $Q_q$ is the empirical quantile of order $q$. The excesses
$Y_k=\{s_{k,i}-u_k \mid s_{k,i}>u_k\}$ are fitted by maximum likelihood to a Generalized Pareto Distribution (GPD) with shape $\xi_k$ and scale $\sigma_k>0$, provided that $n_{u,k}=|Y_k|\geq N_{\min}$. Let $p_{u,k}=n_{u,k}/n_k$ be the empirical probability of exceeding $u_k$. For a target upper-tail probability $\alpha$, the resulting threshold is
\vspace{0.15cm}
\begin{equation}
T_{\mathrm{EVT},k} =
\begin{cases}
u_k +
\dfrac{\sigma_k}{\xi_k}
\left[
\left(
\dfrac{p_{u,k}}{\alpha}
\right)^{\xi_k}
-1
\right],
& |\xi_k|>\varepsilon, \\[8pt]
u_k +
\sigma_k
\log\left(
\dfrac{p_{u,k}}{\alpha}
\right),
& |\xi_k|\leq\varepsilon,
\end{cases}
\label{eq:evt-threshold}
\end{equation}
\vspace{0.15cm}
where $\varepsilon$ handles the limiting case $\xi_k\approx0$. If fewer than $N_{\min}$ excesses are available or the GPD fit is invalid, the empirical quantile $Q_{1-\alpha}(S_k)$ is used as fallback.

To prevent abrupt variations between consecutive windows, we first compute the
exponentially smoothed threshold
$\widetilde{T}_k=\lambda T_{\mathrm{EVT},k}+(1-\lambda)T_{k-1}$.
Its relative change is then limited to $\delta$:
\vspace{0.15cm}
\begin{equation}
T_k =
\operatorname{clip}\!\left(
\widetilde{T}_k,\,
(1-\delta)T_{k-1},\,
(1+\delta)T_{k-1}
\right).
\label{eq:threshold-smoothing}
\end{equation}
\vspace{0.15cm}
A score is classified as anomalous when $s_{k,t}>T_k$. All hyperparameters were fixed a priori without label-based optimization: $q=0.98$, $\alpha=0.01$, $N_{\min}=30$, and $\lambda=\delta=0.05$. Thus, the GPD models the most extreme $2\%$ of scores, requires at least 30 excesses, and limits threshold changes to $5\%$ per window. Here, $\alpha$ is a target upper-tail exceedance probability, not a guaranteed \mbox{false-positive rate}.

\subsection{Post-detection Anomaly Grouping}
Following operational practice~\cite{tejedor2025}, detected anomalies separated by less than 6~hours are consolidated into a single event, reducing redundant alarms and improving diagnostic clarity. This threshold reflects expert recommendations from satellite operations centers.

\section{Results}
Table~\ref{tab:metric_results} summarizes the global event-wise performance for both missions. In M1, the framework achieves high precision with moderate recall, yielding $F_{0.5}=0.700$. This conservative behavior is operationally desirable: the framework flags fewer events but with high confidence, minimizing unnecessary operator interventions. In M2, the trade-off is more balanced ($F_{0.5}=0.698$), detecting a larger proportion of anomalies at the cost of lower precision. 

\begin{figure}[htb]
    \centering
        \includegraphics[width=\linewidth]{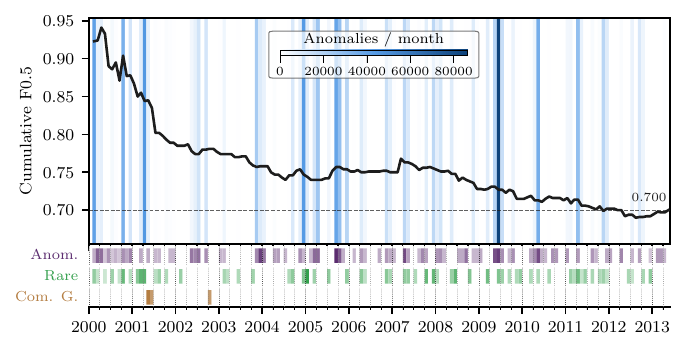}
	\caption{Cumulative corrected $F_{0.5}$ over time for Mission~1. The labels Anom., Rare, and Com.~G. denote anomalous events, rare nominal events, and communication gaps, respectively.}
	\label{fig:acum_f05_m1}
\end{figure}

\begin{figure}[htb]
    \centering
        \includegraphics[width=\linewidth]{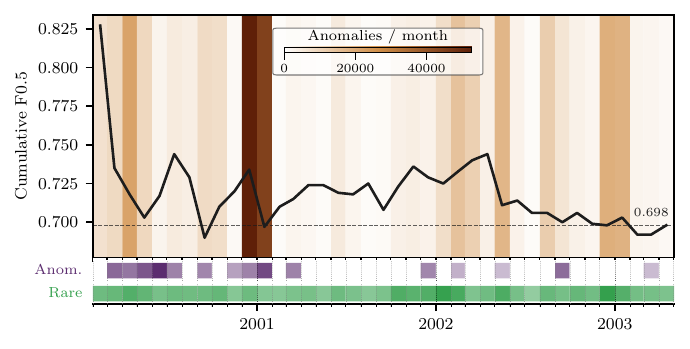}
	\caption{Cumulative corrected $F_{0.5}$ over time for Mission~2. The labels Anom., Rare, and Com.~G. denote anomalous events, rare nominal events, and communication gaps, respectively.}
	\label{fig:acum_f05_m2}
\end{figure}

\begin{table}[htb]
    \begin{center}
    \small
    \begin{tabular}{lcc} 
    \toprule
    \textbf{Metric} & \textbf{Mission 1 (M1)} & \textbf{Mission 2 (M2)}\\
    \midrule
    Event-wise Precision & 0.852 & 0.721 \\
    Event-wise Recall & 0.408 & 0.621 \\
    Event-wise $F_{0.5}$ & 0.700 & 0.698 \\
    True Negative Rate & 0.977 & 0.801 \\
    \bottomrule
    \end{tabular}
    \caption{Global event-wise performance for M1 and M2.}
    \label{tab:metric_results}
    \end{center}
\end{table}
\vspace{-0.35cm}
Figs.~\ref{fig:acum_f05_m1} and~\ref{fig:acum_f05_m2} show the temporal evolution of the cumulative corrected $F_{0.5}$. In M1, anomalous and rare events concentrate heavily in 2000, predominantly in event class~22~\cite[Fig 1]{ESA-ADB-Kotowski2025}, and become sparser afterwards. The $F_{0.5}$ curve reflects this: scores above 0.9 during the initial dense period, gradually stabilizing between 0.7 and 0.8 as anomalies disperse across classes. In M2, event classes are more constant throughout the mission. The cumulative $F_{0.5}$ drops briefly during the first months but recovers quickly, stabilizing around 0.7 with narrower oscillations.

\vspace{-0.5cm}


\paragraph{Comparison with existing baselines.}
Within the ESA-AD ecosystem, industrial unsupervised solutions such as \cite{tejedor2025} have reported event-wise $F_{0.5}$ scores of 0.424 for Mission 1 and 0.882 for Mission 2 under static train/test configurations. In comparison, the proposed framework yields substantially improved results in Mission 1 while maintaining competitive performance in Mission 2. Importantly, these results are obtained under a strictly incremental monthly retraining scheme from the first available period, without reliance on extended historical partitions or mission-specific hyperparameter tailoring, and within a fully deployment-oriented configuration.

For a broader context, \cite{ESA-ADB-Kotowski2025} reports results under a static 50/50 train/test split. Using all channels, the best models achieve $F_{0.5}=0.061$ for M1 and $F_{0.5}=0.241$ for M2. Supervised approaches with channel-subset selection reach $F_{0.5}=0.786$ and $F_{0.5}=0.949$, respectively. Although not directly comparable, our framework achieves $F_{0.5}=0.700$ and $F_{0.5}=0.698$ without any label or channel selection, substantially narrowing the gap with supervised methods. This is particularly significant for M1, where existing unsupervised baselines \mbox{struggle~most.}

\paragraph{Computational Cost.}
Under identical hardware constraints, using CPU-only execution and 16\,GB of RAM, M1 evaluation over 161 monthly windows required 27.5 epochs/window on average, with a mean retraining time of 1041.71\,s/window (total: 46.64\,h) and peak memory averaging 7.72\,GB (max: 8.11\,GB). M2 evaluation over 40 monthly windows required 214.65\,s/window (total: 2.5\,h) and peak memory averaging 0.54\,GB (max: 1.09\,GB). No inter-window memory accumulation was observed in either mission.

\section{Discussion}
Although the framework is evaluated under a strictly incremental regime, the exploratory analysis in Section \ref{3.4: Mission characterization} is performed on the complete mission datasets exclusively for descriptive purposes, to evidence global statistical differences and justify the choice of different model families. It is not used for hyperparameter tuning, threshold calibration, or any label-dependent decision. In operational scenarios, a short warm-up phase with generic detectors can be used to accumulate historical data, after which a statistical assessment restricted to the observed period may guide model selection without introducing information leakage. This separation between statistical characterization and operational training preserves chronological validity while retaining the benefits of structure-aware model selection.

Hyperparameters were selected using standard practices rather than supervised optimization, consistent with the fully unsupervised deployment scenario. When labeled historical data are available, the framework could be extended using automated machine learning (AutoML) strategies \cite{AutoML-Hutter2019} to jointly optimize model architectures, hyperparameters, and thresholds via systematic search procedures guided by event-wise $F_{0.5}$. This would enable adaptive, reproducible model selection across missions, reducing manual-tuning bias.

\begin{acknowledgments}

This work was carried out within the framework of the INCIBE project \textit{AI- and ML-Based Behavioural Anomaly Detection Module}, ref. 250120UCTR-INCIBE (SIMD-I3A Laboratory), with additional support through technical assistance contracts funded by GMV Soluciones Globales Internet, S.A.

This work was also partially funded by the University of Castilla-La Mancha and the European Regional Development Fund (ERDF), ''A Way of Making Europe'', under project 2025-GRIN-38476.
\end{acknowledgments}

\printbibliography
\addcontentsline{toc}{section}{References}

\end{document}